\documentclass[sigconf]{acmart}

\usepackage{booktabs}
\usepackage{graphicx}
\usepackage[tight,footnotesize]{subfigure}
\usepackage{soul}
\usepackage[utf8]{inputenc}
\usepackage{multirow,makecell,footmisc,url}
\usepackage{amsmath,amssymb,amsthm,bm}
\usepackage[ruled]{algorithm2e}
\usepackage{marvosym}

\copyrightyear{2018} 
\acmYear{2018} 
\setcopyright{acmcopyright}
\acmConference[MM '18]{2018 ACM Multimedia Conference}{October 22--26, 2018}{Seoul, Republic of Korea}
\acmBooktitle{2018 ACM Multimedia Conference (MM '18), October 22--26, 2018, Seoul, Republic of Korea}
\acmPrice{15.00}
\acmDOI{10.1145/3240508.3240516}
\acmISBN{978-1-4503-5665-7/18/10}

\begin{document}

\title{Deep Priority Hashing}

\author{Zhangjie Cao, Ziping Sun, Mingsheng Long(\Letter), Jianmin Wang, and Philip S. Yu}
\affiliation{%
\institution{
    School of Software, Tsinghua University, Beijing 100084, China \\
    Beijing National Research Center for Information Science and Technology}
}
\email{{caozhangjie14,sunziping2016}@gmail.com, {mingsheng,jimwang}@tsinghua.edu.cn, psyu@uic.edu}

\renewcommand{\shortauthors}{Cao et al.}
\graphicspath{{figures/}}

\begin{abstract}
Deep hashing enables image retrieval by end-to-end learning of deep representations and hash codes from training data with pairwise similarity information. Subject to the distribution skewness underlying the similarity information, most existing deep hashing methods may underperform for  imbalanced data due to misspecified loss functions. This paper presents Deep Priority Hashing (DPH), an end-to-end architecture that generates compact and balanced hash codes in a Bayesian learning framework. The main idea is to reshape the standard cross-entropy loss for similarity-preserving learning such that it down-weighs the loss associated to highly-confident pairs. This idea leads to a novel \emph{priority} cross-entropy loss, which prioritizes the training on uncertain pairs over confident pairs. Also, we propose another \emph{priority} quantization loss, which prioritizes hard-to-quantize examples  for generation of nearly lossless hash codes. Extensive experiments demonstrate that DPH can generate high-quality hash codes and yield state-of-the-art image retrieval results on three datasets, ImageNet, NUS-WIDE, and MS-COCO.
\end{abstract}




\maketitle

\section{Introduction}

Multimedia data has been ubiquitous in search engines and online communities, while its efficient retrieval is important to enhance user experience. The major challenges in multimedia retrieval reside in the large-scale and high-dimension of multimedia data.
To enable accurate retrieval under efficient computation, approximate nearest neighbors (ANN) search has attracted increasing attention. Parallel to the traditional indexing methods \cite{cite:TOMM06CBIR} for candidates pruning, another advantageous solution is hashing methods \cite{cite:TPAMI2018HashSurvey} for data compression, which transform high-dimensional media data into compact binary codes while similar binary codes are generated for similar data items. In this paper, we focus on the learning to hash methods \cite{cite:TPAMI2018HashSurvey}, which build data-dependent hash encoding schemes for efficient image retrieval. These methods can capture the underlying data distributions to achieve better performance than traditional data-independent hashing methods, e.g. Locality-Sensitive Hashing (LSH) \cite{cite:VLDB99LSH}.

A fruitful line of learning to hash methods have been designed to enable efficient ANN search, where the efficiency comes from the compact binary codes that are orders of magnitude smaller than the original high-dimensional feature descriptors. Ranking these binary codes in response to each query entails only a few computations of the Hamming distance between the query and each database item. Hash lookup further reduces the search to constant time by early pruning of irrelevant candidates falling out of a small Hamming ball. The literature can be divided into supervised and unsupervised paradigms \cite{cite:NIPS09BRE,cite:CVPR11ITQ,cite:ICML11MLH,cite:CVPR12MIH,cite:CVPR12KSH,cite:TPAMI12SSH,cite:CVPR13HBS,cite:CVPR14CH,cite:SIGIR14LFH}.
Recently, deep learning to hash methods  \cite{cite:AAAI14CNNH,cite:CVPR15DNNH,cite:CVPR15SDH,cite:CVPR15DH,cite:AAAI16DHN,cite:IJCAI16DPSH,cite:CVPR2016DSH,cite:ICCV17HashNet} have shown that deep neural networks can be used as nonlinear hash functions to enable end-to-end learning of deep representations and hash codes. These deep hashing methods have shown state-of-the-art results. In particular, it proves crucial to jointly learn similarity-preserving representations and control quantization error of converting continuous representations to binary codes \cite{cite:AAAI16DHN,cite:IJCAI16DPSH,cite:CVPR2016DSH,cite:ICCV17HashNet}. 

Most of the existing methods are tailored to image retrieval scenarios with balanced or nearly balanced data. In other words, they weigh equally each data pair, no matter they are similar pairs or dissimilar pairs. Thus, they can maximize retrieval performance on average per-instance accuracy. However, due to the well-known long-tail law, multimedia data with skewed distribution are prevalent in many online image search systems. The data skewness may either stem from the imbalanced numbers of similar and dissimilar pairs associated with each query, or from the diversity of popular and rare classes, or even from the variations in easy and difficult pairs of images. Such data skewness will severely affect the retrieval performance, especially when one needs to trade off the precision (weigh dissimilar pairs more in order to discard irrelevant results) from recall (weigh similar pairs more in order to include potentially relevant results).
Therefore, how to address various data skewness problems simultaneously remains an open problem.

This work presents Deep Priority Hashing (DPH), a novel deep hashing model that generates compact binary codes to enable effective and efficient image retrieval under data skewness problems. DPH is formalized as a Bayesian learning framework, providing two novel loss functions motivated by the success of the focal loss in object detection problem~\cite{cite:ICCV17Focal}. One is a \emph{priority} cross-entropy loss for similarity-preserving learning, which prioritizes difficult image pairs over easy image pairs to learn prioritized deep representations. The other is a \emph{priority} quantization loss, which prioritizes hard-to-quantize examples for generating nearly lossless hash codes. Both loss functions are well-specified to similarity retrieval of highly skew image data. The proposed DPH model is an end-to-end architecture that can be trained by standard back-propagation.
Extensive experiments demonstrate that DPH can generate high-quality hash codes and yield state-of-the-art image retrieval performance on three benchmark datasets, ImageNet, NUS-WIDE, and MS-COCO.

\section{Related Work}

Learning to hash has become an important research direction in multimedia retrieval, which trades off efficacy from efficiency. Wang et al. \cite{cite:TPAMI2018HashSurvey} has provided a comprehensive literature survey that covers most of important methods and latest advances.

Existing hashing methods can be divided into unsupervised hashing and supervised hashing.
Unsupervised hashing methods learn hash functions that encode data points to binary codes by training solely from unlabeled data. Typical learning criteria include reconstruction error minimization \cite{cite:AI07SemanticHashing,cite:CVPR11ITQ,cite:TPAMI11PQ} and graph structure preservation \cite{cite:NIPS09SH,cite:ICML11AGH}. While unsupervised methods are more general and can be trained without semantic labels or relevance information, they are subject to the semantic gap dilemma \cite{cite:TPAMI00SemanticGap} that high-level semantic description of an object differs from low-level feature descriptors. Supervised methods can incorporate semantic labels or relevance information to mitigate the semantic gap and improve the hashing quality. 
Typical supervised methods include Binary Reconstruction Embedding (BRE) \cite{cite:NIPS09BRE}, Minimal Loss Hashing (MLH) \cite{cite:ICML11MLH}, Hamming Distance Metric Learning \cite{cite:NIPS12HDML}, and Supervised Hashing with Kernels (KSH) \cite{cite:CVPR12KSH}, which generate hash codes by minimizing the Hamming distances across similar pairs and maximizing the Hamming distances across dissimilar pairs.

As deep convolutional networks \cite{cite:NIPS12CNN,cite:CVPR16DRL} yield sharp performance on many computer vision tasks, deep learning to hash has attracted attention recently. CNNH \cite{cite:AAAI14CNNH} adopts a two-stage strategy in which the first stage learns hash codes and the second stage learns a deep network to map input images to the hash codes. DNNH \cite{cite:CVPR15DNNH} improved the two-stage CNNH with a simultaneous feature learning and hash coding  pipeline such that representations and hash codes can be optimized in a joint optimization process. DHN \cite{cite:AAAI16DHN} further improves DNNH by a cross-entropy loss and a quantization loss which preserve the pairwise similarity and control the quantization error simultaneously. DHN obtains state-of-the-art performance on several benchmarks. DPSH \cite{cite:IJCAI16DPSH} and DSH \cite{cite:CVPR2016DSH} follow similar framework as DHN, thus yielding similar retrieval performance. 
HashNet \cite{cite:ICCV17HashNet} improves DHN by balancing the positive and negative pairs in training data to trade off precision vs. recall, and by the continuation technique to yield exactly binary codes with the lowest quantization error. HashNet obtains state-of-the-art performance on several benchmark datasets.

However, existing deep hashing methods do not consider the data skewness problem. In other words, they let all the training pairs contribute equally to the loss function, where easy pairs will overwhelm the loss function so that the difficult ones cannot be trained sufficiently. To address these problems, we propose a novel Deep Priority Hashing (DPH) model. We design a novel priority cross-entropy loss to concentrate on difficult pairs more than on easy pairs. We design another priority quantization loss to concentrate on hard-to-quantize examples for generating less lossy hash codes. This work is among the earliest endeavors on deep hashing with the \emph{prioritization} towards different data skewness scenarios. 

\begin{figure*}[tbp]
  \centering
  \includegraphics[width=1.0\textwidth]{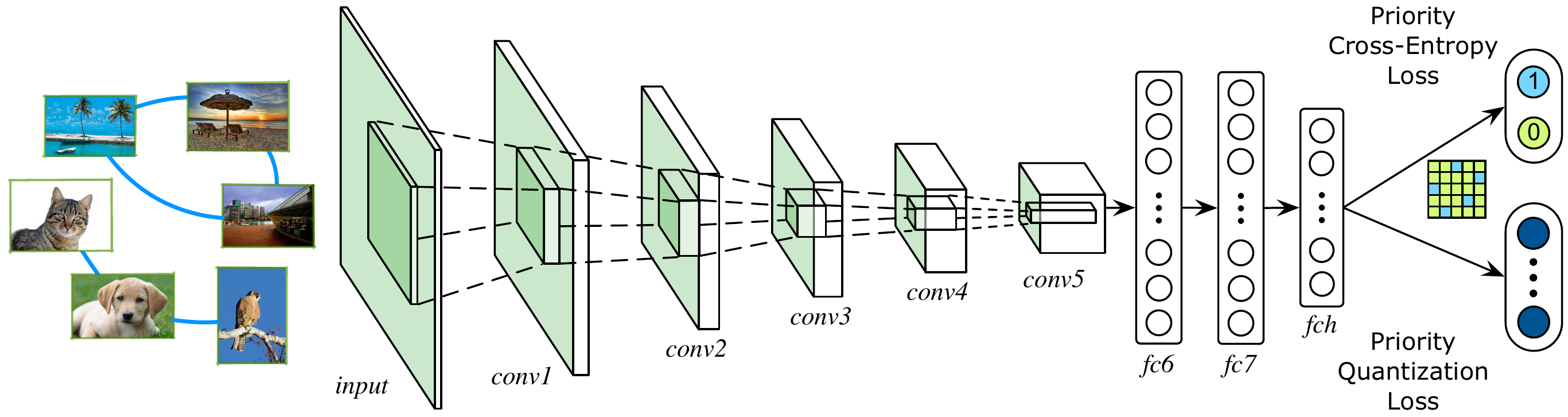}
  \caption{The architecture of Deep Priority Hashing (DPH) consists of four components: {1)} a convolutional network (CNN) for learning deep representation of each point ${\bm x}_i$, {2)} a fully-connected hash layer (\textit{fch}) for transforming the deep representation into $K$-bit hash code ${\bm h}_i$, {3)} a priority cross-entropy loss that prioritizes difficult pairs over easy pairs for similarity-preserving learning, and {4)} a priority quantization loss that prioritizes hard-to-quantize points for controlling the hashing quality.}
  \label{fig:DPH}
\end{figure*}

\section{Preliminary on Focal Loss}

The Focal Loss was introduced by He et al. \cite{cite:ICCV17Focal}, which yields state-of-the-art performance for object detection. It is designed to address the problem of an extreme imbalance between examples of different classes (e.g. foreground and background classes) during training. The Focal Loss has a close connection to the cross-entropy loss for binary classification. The Cross-Entropy (CE) loss is defined as
\begin{equation}\label{eqn:CE}
\begin{aligned}
{\text{CE}(p,y)} &=
  \begin{cases}
    -\log(p) & \quad \text{if } {y} = 1 \\
    -\log(1-p) &\quad \text{otherwise}. \\
  \end{cases}
  \end{aligned}
\end{equation}
In the above $y \in \{ \pm 1\}$ specifies the ground-truth class and
$p \in [0, 1]$ is the model's estimated probability for the class of label $y = 1$. For  notational convenience, He et al. \cite{cite:ICCV17Focal} defined $p_t$ as
\begin{equation}\label{eqn:pt}
\begin{aligned}
{p_t} &=
  \begin{cases}
    p & \quad \text{if } {y} = 1 \\
    1-p & \quad \text{otherwise}. \\
  \end{cases}
    \end{aligned}
\end{equation}
and rewrote $\text{CE}(p, y) = \text{CE}(p_t) = - \log(p_t)$.

One notable property of the cross-entropy loss is that even for easily classified examples ($p_t \gg 0.5$) there incurs a loss with non-trivial magnitude. When summed over a large number of easy examples, these small loss values can accumulate to overwhelm the rare class with difficult examples. Towards this problem, He et al. \cite{cite:ICCV17Focal} proposes to reshape the cross-entropy loss into Focal Loss, which down-weighs easy examples and focuses on difficult examples.

The Focal Loss (FL) adds a modulating factor $(1-p_t)^\gamma$ to the cross-entropy loss with a tunable focusing parameter $\gamma \ge 0$:
\begin{equation}\label{eqn:focal_loss}
\text{FL} \left(p_t \right) = -\left(1- p_t \right)^\gamma \log \left(p_t \right).
\end{equation}
Focal Loss has two nice properties. 1) When an example is misclassified and $p_t$ is small, the modulating factor is near $1$ and the loss is unaffected. As $p_t \rightarrow 1$, the factor goes to $0$ and the loss for well-classified examples is down-weighed. 2) The focusing parameter $\gamma$ smoothly adjusts the rate at which easy examples are down-weighed. When $\gamma = 0$, $\text{FL}$ is equivalent to $\text{CE}$, and as $\gamma$ increases the effect of the modulating factor is likewise increase.

Intuitively, the modulating factor reduces the contribution from easy examples to the loss and enlarges the loss gap between easy and difficult examples. This in turn increases the force to correct misclassified examples.

In practice, we prefer an $\alpha$-balanced variant of the focal loss:
\begin{equation}\label{eqn:alpha_focal_loss}
\text{FL} \left(p_t \right) = -{\alpha}_t \left(1- p_t \right)^\gamma \operatorname{log}\left(p_t \right).
\end{equation}
This variant of the focal loss can further address the class imbalance problem between large classes and rare classes, which is a typical option for practical applications.

A limitation of the focal loss is that the modulating factor $(1-p_t)^\gamma$ is strictly determined by classification uncertainty $p_t$. In many real scenarios, we may need to consider other measures to quantify the \emph{difficulty} of image pairs. The flexibility in choosing different modulating factors for prioritization is the motivation of our work.

\section{Deep Priority Hashing}

In similarity retrieval, we are given a training set of $N$ points $\{{\bm x}_i\}_{i=1}^N$, each represented by a $D$-dimensional feature vector ${\bm x}_i \in \mathbb{R}^{D}$. Some pairs of points ${\bm x}_i$ and ${\bm x}_j$ are provided with similarity labels $s_{ij}$, where $s_{ij} = 1$ if ${\bm x}_i$ and ${\bm x}_j$ are similar while $s_{ij} = 0$ if ${\bm x}_i$ and ${\bm x}_j$ are dissimilar. 
Deep hashing learns a nonlinear hash function $f:{\bm{x}} \mapsto {\bm{h}} \in {\left\{ { - 1,1} \right\}^K}$ from input space $\mathbb{R}^D$ to Hamming space $\{-1,1\}^K$ with deep network. It encodes each point ${\bm x}$ into $K$-bit hash code ${\bm h} = f({\bm x})$ such that the similarity in the training pairs $\{({\bm x}_i, {\bm x}_j, s_{ij}): s_{ij} \in \mathcal{S}\}$ can be preserved in the Hamming space. The similarity information $\mathcal{S} = \{s_{ij}\}$ can be collected from semantic labels or relevance feedback in online search systems.

This paper presents Deep Priority Hashing (\textbf{DPH}), an end-to-end architecture to enable efficient image retrieval, as shown in Figure~\ref{fig:DPH}. 
The proposed deep architecture accepts pairwise input images $\{({\bm x}_i, {\bm x}_j, s_{ij})\}$ and processes them through a pipeline of deep representation learning and binary hash coding: {1)} a convolutional network (CNN) for learning deep representation of each image ${\bm x}_i$, {2)} a fully-connected hash layer (\textit{fch}) for transforming the deep representation into $K$-bit hash code ${\bm h}_i \in \{1,-1\}^K$, {3)} a priority cross-entropy loss that prioritizes difficult pairs over easy pairs for similarity-preserving learning, and {4)} a priority quantization loss that prioritizes hard-to-quantize images for controlling the binarization error due to continuous relaxation in the optimization.

\subsection{Deep Architecture}

Figure~\ref{fig:DPH} illustrates the architecture of the proposed Deep Priority Hashing. We extend from AlexNet \cite{cite:NIPS12CNN}, a deep convolutional neural network (CNN) with five convolutional layers \textit{conv1}--\textit{conv5} and three fully-connected layers \textit{fc6}--\textit{fc8}. We replace the classifier layer \textit{fc8} with a new hash layer \textit{fch} of $K$ hidden units, which transforms the representation of the \textit{fc7} layer into $K$-dimensional continuous code  ${\bm z}_i \in \mathbb{R}^K$. We can obtain hash code ${\bm h}_i$ through the sign thresholding ${\bm h}_{i} = \operatorname{sgn} ({\bm z}_i)$. However, we adopt the hyperbolic tangent (tanh) function to squash the continuous code to be within $[-1,1]$ instead of using sign function. To further guarantee the quality of hash codes for efficient image retrieval, we preserve the similarity in training pairs $\{({\bm x}_i, {\bm x}_j, s_{ij}): s_{ij} \in \mathcal{S}\}$ by designing a \emph{priority} cross-entropy loss and control the quantization error by designing another \emph{priority} quantization loss. Both loss functions can be derived in the Maximum a Posteriori (MAP) estimation framework. Though following most work to use AlexNet in Figure~\ref{fig:DPH}, we can easily replace the backbone network in our architecture with any classification network since we only replace the last classifier layer while the other layers can be inherited from the backbone network.

\subsection{Model Formulation}

This paper enables deep hashing from skew data with both easy and difficult examples by a Bayesian learning framework. The framework jointly preserves similarity information of pairwise images and controls the quantization error of continuous relaxation. For a pair of hash codes ${\bm h}_i$ and ${\bm h}_j$, their Hamming distance $\mathrm{d}_H(\cdot,\cdot)$ and their inner product $\langle \cdot,\cdot \rangle$ satisfy ${\textrm{d}}_H\left( {{{\bm{h}}_i},{{\bm{h}}_j}} \right) = \frac{1}{2}\left( {K - \left\langle {{{\bm{h}}_i},{{\bm{h}}_j}} \right\rangle } \right)$, indicating that the Hamming distance and inner product can be used interchangeably for binary codes. Thus we adopt inner product to quantify the pairwise similarity. 
Given training image pairs with pairwise similarity labels as $\{({\bm x}_i, {\bm x}_j, s_{ij}): s_{ij} \in \mathcal{S}\}$, the logarithm Weighted Maximum a Posteriori (WMAP) estimation of the hash codes ${\bm H} = [{\bm h}_1,\ldots,{\bm h}_N]$ for $N$ training images can be defined as
\begin{equation}\label{eqn:WMAP}
\begin{aligned}
\log p\left( {{\bm{H}}|{\mathcal{S}}} \right) & \propto \log p\left( {{\mathcal{S}}|{\bm{H}}} \right) p\left( {\bm{H}} \right)\\
  & = \sum\limits_{{s_{ij}} \in {\mathcal{S}}} {{w_{ij}}\log p\left( {{s_{ij}}|{{\bm{h}}_i},{{\bm{h}}_j}} \right)} + \;\, \sum\limits_{i = 1}^{N} {w'_i} {\log p\left( {{{\bm{h}}_i}} \right)},
\end{aligned}
\end{equation}
where $p\left( {\mathcal{S}|{\bm{H}}} \right) = \prod\nolimits_{{s_{ij}} \in \mathcal{S}} {{{\left[ {p\left( {{s_{ij}}|{{\bm{h}}_i},{{\bm{h}}_j}} \right)} \right]}^{{w_{ij}}}}} $ is the weighted likelihood function for pairwise data, and $w_{ij}$ is the weight for each training pair $({\bm x}_i, {\bm x}_j, s_{ij})$. This is extended from the weighted maximum likelihood on pointwise data \cite{cite:JMLR10MWL}. Another difference from \cite{cite:JMLR10MWL} is the weighted prior $p\left( {\bm{H}} \right)$, and ${w'_i}$ is the weight for an image ${\bm x}_i$. In this paper, we propose the above Weighted Maximum a Posteriori (WMAP) estimation over pairwise data with different skewness scenarios. This is a general framework for instantiating specific learning to hash models, by choosing well-specified probability functions and weighting schemes.

\subsubsection{Priority Cross-Entropy Loss}

We first derive the priority cross-entropy loss for similarity-preserving learning.
For each image pair with label $({\bm x}_i, {\bm x}_j, s_{ij})$, $p(s_{ij}|{\bm h}_i,{\bm h}_j)$ is the conditional probability of the similarity label $s_{ij}$ given the pair of corresponding hash codes ${\bm h}_i$ and ${\bm h}_j$, which is defined by the logistic function:
\begin{equation}\label{eqn:Pij}
	\begin{aligned}
	p\left( {{s_{ij}}|{{\bm{h}}_i},{{\bm{h}}_j}} \right) &= 
		\begin{cases}
			\sigma \left( {\left\langle {{\bm{h}}_i, {\bm{h}}_j} \right\rangle } \right), & {s_{ij}} = 1 \\
			1 - \sigma \left( {\left\langle {{\bm{h}}_i, {\bm{h}}_j} \right\rangle } \right), & {s_{ij}} = 0 \\
		\end{cases} \\
		& = \sigma {\left( {\left\langle {{\bm{h}}_i,{\bm{h}}_j} \right\rangle } \right)^{{s_{ij}}}}{\left( {1 - \sigma \left( {\left\langle {{\bm{h}}_i,{\bm{h}}_j} \right\rangle } \right)} \right)^{1 - {s_{ij}}}} \\
	\end{aligned}
\end{equation}
where $\sigma \left( x \right) = {1}/({{1 + {e^{ - \beta x}}}})$ is an {adaptive} variant of the sigmoid function with parameter $\beta$ to control its bandwidth. As the sigmoid function with larger $\beta$ has larger saturation zone, we usually require $\beta < 1$ to perform back-propagation with more gradients.

Motivated by the Focal Loss (FL) \cite{cite:ICCV17Focal}, we use the weight combined by $\alpha_t$-scaling and modulating factor $(1-q)^\gamma$ to simultaneously model the class diversity (including imbalance in both similar and dissimilar pairs) and the variation in easy and difficult examples. But unlike the focal loss, in the modulating factor we use a different measure $q$ instead of the classification uncertainty $p$ to quantify the \emph{difficulty} of each image pair. This new design relaxes the restriction in the focal loss that the ``difficulty'' of each image pair and the classification uncertainty should be consistent. In this way, we can adopt more flexible choices for the modulating factor. Furthermore, in deep hashing we do not have images with pointwise class labels, and thus need to model image pairs with pairwise labels $({\bm x}_i, {\bm x}_j, s_{ij})$. We propose a novel \emph{priority} weighting scheme as follows,
\begin{equation}\label{eqn:Wij}
	{w_{ij}} = \alpha_{ij} \left(1-q_{ij} \right)^\gamma,
\end{equation}
where $w_{ij}$ is the weight for each training pair $({\bm x}_i, {\bm x}_j, s_{ij})$. It consists of a scaling part to weigh for class imbalance problem, and a modulating part to weigh for easy and hard examples. Specifically, the measure of \emph{difficulty} for the modulating part, i.e. $q_{ij} = q \left( {{s_{ij}}|{{\bm{h}}_i},{{\bm{h}}_j}} \right)$ is defined as 
\begin{equation}\label{eqn:Qij}
	\begin{aligned}
	q \left( {{s_{ij}}|{{\bm{h}}_i},{{\bm{h}}_j}} \right) &= 
		\begin{cases}
			{\frac{{1 + \cos \left( {{{\bm{h}}_i},{{\bm{h}}_j}} \right)}}{2}}, & {s_{ij}} = 1 \\
			{\frac{{1 - \cos \left( {{{\bm{h}}_i},{{\bm{h}}_j}} \right)}}{2}}, & {s_{ij}} = 0 \\
		\end{cases} \\
		& = {\left( {\frac{{1 + \cos \left( {{{\bm{h}}_i},{{\bm{h}}_j}} \right)}}{2}} \right)^{{s_{ij}}}}{\left( {\frac{{1 - \cos \left( {{{\bm{h}}_i},{{\bm{h}}_j}} \right)}}{2}} \right)^{1 - {s_{ij}}}} \\
	\end{aligned}
\end{equation}
where $q_{ij}$ indicates how difficult an image pair $({\bm x}_i, {\bm x}_j)$ is classified as similar when $s_{ij}=1$, or classified as dissimilar when $s_{ij}=0$. With these difficulty quantities $q_{ij}$, we can put different weights on easy and hard examples to prioritize on more difficult image pairs. An important motivation of using $q_{ij}$ as the measure of ``difficulty'' is that $q_{ij}$ is \emph{magnitude-invariant}, consistent with the fact that hash codes have the same magnitude for different images. Therefore, $q_{ij}$ can potentially eliminate the variations in code magnitudes due to skew data distributions.

In the original focal loss, the scaling part tackles the imbalance between different classes in a pointwise scenario, which is not applicable to pairwise scenario. In HashNet \cite{cite:ICCV17HashNet}, the number of similar pairs and dissimilar pairs are used to compute the weight for data imbalance. However, such a strategy cannot quantify the imbalance of underlying different classes, since the class information is unavailable in pairwise scenarios. Quantification of the scaling part for the pairwise scenario remains a nontrivial problem unsolved by previous work. In this paper, we define the scaling part $\alpha_{ij}$ by considering the degrees of each vertex in the similarity graph ${\mathcal S}$ as
\begin{equation}\label{eqn:Aij}
{\alpha_{ij}} = 
  \begin{cases}
    \frac{{\left| {{\mathcal{S}_i}} \right|\left| {{\mathcal{S}_j}} \right|}}{{\sqrt {\left| {\mathcal{S}_i^1} \right|\left| {\mathcal{S}_j^1} \right|} }}, & {s_{ij}} = 1 \\
    \frac{{\left| {{\mathcal{S}_i}} \right|\left| {{\mathcal{S}_j}} \right|}}{{\sqrt {\left| {\mathcal{S}_i^0} \right|\left| {\mathcal{S}_j^0} \right|} }}, & {s_{ij}} = 0 \\ 
  \end{cases}
\end{equation}
where $\alpha_{ij}$ indicates the pairwise weight characterizing the class imbalance, which is influenced by both images in that pair.
${\mathcal{S}_i} = \left\{ {{s_{ij}} \in \mathcal{S}:\forall j} \right\}$ is the set of pairs containing ${\bm x}_i$. It is divided into two subsets: ${\mathcal{S}^1_i} = \left\{ {{s_{ij}} \in \mathcal{S}:\forall j,{s_{ij}} = 1} \right\}$ is the subset of similar pairs that contain ${\bm x}_i$, while ${\mathcal{S}^0_i} = \left\{ {{s_{ij}} \in \mathcal{S}:\forall j,{s_{ij}} = 0} \right\}$ is the subset of dissimilar pairs that contain ${\bm x}_i$. $\mathcal{S}_j$, ${\mathcal{S}^1_j}$ and ${\mathcal{S}^0_j}$ are similarly defined. Intuitively, for each image ${\bm x}_i$, ${\left| {\mathcal{S}_i^1} \right|}$ and ${\left| {\mathcal{S}_i^0} \right|}$ respectively calculate the numbers of its similar and dissimilar images, which naturally captures the data skewness due to the class imbalance.

By taking the priority weight in Equation~\eqref{eqn:Wij} into the WMAP estimation in Equation~\eqref{eqn:WMAP} and letting $p_{ij} = p \left( {{s_{ij}}|{{\bm{h}}_i},{{\bm{h}}_j}} \right)$, we can obtain a novel (pairwise) Priority Cross-Entropy Loss as follows:
\begin{equation}\label{eqn:L}
	L =  - \sum\limits_{{s_{ij}} \in \mathcal{S}} {\alpha _{ij}{{\left( {1 - q_{ij}} \right)}^\gamma }\log \left( {p_{ij}} \right)}.
\end{equation}
The priority cross-entropy loss is a natural extension of the focal loss to the pairwise classification scenario, which inherits all nice properties of the focal loss and broadens the choices of the modulating factor to incorporate more reasonable measures of ``difficulty''. More specifically, the priority cross-entropy loss will down-weigh confident pairs and prioritize on difficult pairs with low confidence. And with the proposed scaling strategy in Equation~\eqref{eqn:Aij}, the priority cross-entropy loss assigns larger weights to rare classes and smaller weights to popular classes, which for the first time, addresses the class imbalance issue in pairwise scenario for image retrieval.

\subsubsection{Priority Quantization Loss}

To control the quantization error of the binarization operation, we further derive the priority quantization loss from the MAP framework to yield nearly lossless hash codes.
As discrete optimization of Equation~\eqref{eqn:WMAP} with binary constraints ${\bm h}_i\in\{-1,1\}^K$ is very challenging, \emph{continuous relaxation} widely adopted by existing hashing methods \cite{cite:TPAMI2018HashSurvey} to apply to the binary constraints for ease of optimization. However, the continuous relaxation will give rise to two important technical issues: 1) uncontrollable quantization error caused by binarizing continuous codes to binary codes and 2) large approximation error by adopting inner product between continuous codes as the surrogate of Hamming distance between binary codes. To control the quantization error and close the gap between Hamming distance and its surrogate, in this paper, we propose an (unnormalized) bimodal Laplacian prior for the continuous codes $\{{\bm h}_i\}$, which is defined as:
\begin{equation}\label{eqn:prior}
	p\left( {\bm{h}}_i \right) = \frac{1}{{2\epsilon}}\exp \left( { - \frac{{{{\left\| {\left| {\bm{h}}_i \right| - {\bm 1} } \right\|}_1}}}{\epsilon}} \right) ,
\end{equation}
where $\epsilon$ is the diversity parameter, and let $p_i = p\left( {\bm{h}}_i \right)$. We can validate that the prior puts the largest density on the discrete values $\{-1,1\}$, which enforces that the learned Hamming embeddings $\{{\bm h}_i\}$ should be assigned to $\{-1,1\}$ with the largest probability.

Similar to the priority cross-entropy loss, we define the priority weight for the priority quantization loss as
\begin{equation}\label{eqn:Wi}
	{{w'_i}} = \alpha _i{\left( {1 - q_i} \right)^\gamma },
\end{equation}
which consists of a scaling part and a modulating part. Note that we need to control the quantization error of all images equally, regardless of whether they are from the rare classes. Thus we set constant scaling $\alpha_i = 1$. The modulating part controls the quantization error under the variations in easy and hard examples, with $q_i$ defined as
\begin{equation}\label{eqn:Pi_focal}
	q_i=q\left( {{{\bm{h}}_i}} \right) = \frac{{1 + \cos \left( {\left| {{{\bm{h}}_i}} \right|,{\bm{1}}} \right)}}{2},
\end{equation}
which indicates how likely a continuous code ${\bm h}_i$ can be perfectly quantized into binary code ${\operatorname{sgn} \left( {{{\bm{h}}_i}} \right)}$.
With probabilities $q_{i}$, we can put different weights on easy-to-quantize and hard-to-quantize examples and prioritize hard-to-quantize ones. 

By taking the priority weight in Equation~\eqref{eqn:Wi} into the WMAP in Equation~\eqref{eqn:WMAP}, we obtain a novel Priority Quantization Loss as:
\begin{equation}\label{eqn:Q}
	Q =  - \sum\limits_{{s_{ij}} \in \mathcal{S}} {{{\left( {1 - q_i} \right)}^\gamma }\log \left( {p_i} \right)}.
\end{equation}
The priority quantization loss helps generating nearly lossless hash codes by reducing the quantization error of hard-to-quantized examples more than of easy-to-quantize examples.

\subsubsection{Hash Function Learning}

This paper establishes deep learning to hash for skew data with pairwise similarity information, which constitutes two key components: Priority Cross-Entropy Loss for similarity-preserving learning and Priority Quantization Loss for generating nearly lossless hash codes.
The overall optimization problem is an integration of the Priority Cross-Entropy Loss in Equations \eqref{eqn:L} and the Priority Quantization Loss in Equation \eqref{eqn:Q}:
\begin{equation}\label{eqn:model}
	\mathop {\min }\limits_\Theta  L + Q,
\end{equation}
where $\Theta $ is the network parameters efficiently optimized using standard back-propagation with automatic differentiation techniques. Note that, the scale parameter $\frac{1}{\epsilon}$ in Equation~\eqref{eqn:prior} introduces a tradeoff hyper-parameter between the two losses.

Based on the WMAP estimation in Equation~\eqref{eqn:model}, we can learn compact hash codes by jointly preserving the pairwise similarity and controlling the quantization error. Finally, we can obtain $K$-bit binary codes by simple sign thresholding ${\bm h} \leftarrow \text{sgn}({\bm h})$, where $\text{sgn}({\bm h})$ is the sign function on vectors that for $i = 1,\ldots,K$, $\text{sgn}(h_i) = 1$ if $h_i > 0$, otherwise $\text{sgn}(h_i) = -1$. It is worth noting that, since we have minimized the quantization error in \eqref{eqn:model} during training, this final binarization step will incur very small loss of retrieval quality.

\section{Experiments}

We conduct extensive experiments to evaluate DPH with several state-of-the-art hashing methods on three benchmark datasets. Codes and datasets are available at \url{https://github.com/thuml/DPH}.

\subsection{Setup}

\textbf{ImageNet} is a benchmark dataset for Large Scale Visual Recognition Challenge (ILSVRC 2015) \cite{cite:ILSVRC15}. It contains over 1.2M images in the training set and 50K images in the validation set, where each image is single-labeled by one of the 1,000 categories. We use the sample of 100 categories organized by HashNet \cite{cite:ICCV17HashNet}. We use the same database set and query set but re-sample the training set to make it data-skew. Our training set contains 10,000 images, which consists of three groups: the first group with $1$ big class ($\sim$1,300 images/class), the second group with $9$ middle classes ($\sim$400 images/class) and the third group with $90$ small classes ($\sim$50 images/class). The classes of each group are randomly sampled. 

\textbf{NUS-WIDE}\footnote{\url{http://lms.comp.nus.edu.sg/research/NUS-WIDE.htm}} \cite{cite:CIVR09NusWide} is an image dataset containing 269,648 images from \url{Flickr.com}. Each image is annotated by some of the 81 ground truth concepts (categories). We follow a slightly different evaluation protocol as HashNet \cite{cite:ICCV17HashNet}. We randomly sample 50 images per class as query images, with the remaining images used as the database. We further sample 10,000 images randomly from the database as training images.

\textbf{MS-COCO}\footnote{\url{http://mscoco.org}} \cite{cite:MSCOCO} is an image recognition, segmentation, and captioning dataset. The current release contains 82,783 training images and 40,504 validation images, where each image is labeled by some of the 80 categories. We obtain the 12,2218 images set by combining the training and validation images. As described above in the NUS-WIDE dataset, we randomly sample 50 images per class as queries. We use the rest images as the database and randomly sample 10,000 images from the database as training images.

Following standard evaluation protocol as previous work \cite{cite:AAAI14CNNH,cite:CVPR15DNNH,cite:AAAI16DHN,cite:ICCV17HashNet}, the similarity information for hash function learning and for ground-truth evaluation is constructed from image labels: if two images $i$ and $j$ share at least one label, they are similar and  $s_{ij}=1$; otherwise, they are dissimilar and $s_{ij}=0$. Note that, although we use the image labels to construct the similarity information, our proposed DPH model can learn hash codes when only the similarity information is available. These datasets exhibit data skewness problems and thus can be used to evaluate different hashing methods in the data skewness scenario.

We compare \textbf{DPH} in terms of the retrieval performance against eleven classical or state-of-the-art hashing methods, including unsupervised hashing methods \textbf{LSH} \cite{cite:VLDB99LSH}, \textbf{SH} \cite{cite:NIPS09SH}, \textbf{ITQ} \cite{cite:CVPR11ITQ}, supervised shallow hashing methods \textbf{KSH} \cite{cite:CVPR12KSH}, \textbf{SDH} \cite{cite:CVPR15SDH}, and supervised deep hashing methods  \textbf{CNNH} \cite{cite:AAAI14CNNH}, \textbf{DNNH} \cite{cite:CVPR15DNNH}, \textbf{DPSH} \cite{cite:IJCAI16DPSH}, \textbf{DSH} \cite{cite:CVPR2016DSH}, \textbf{DHN} \cite{cite:AAAI16DHN}, \textbf{HashNet} \cite{cite:ICCV17HashNet}.

\begin{table*}[tb]
    \centering 
    \addtolength{\tabcolsep}{0pt}
    \caption{Mean Average Precision (MAP) of Hamming Ranking for Different Number of Bits on the Three Image Datasets}
    \label{table:MAP}
    \begin{tabular}{c|cccc|cccc|cccc}
        \Xhline{1.0pt}
        \multirow{2}{30pt}{\centering Method} & \multicolumn{4}{c|}{ImageNet} & \multicolumn{4}{c|}{NUS-WIDE} & \multicolumn{4}{c}{MS-COCO} \\
        \cline{2-13}
        & 16 bits & 32 bits  & 48 bits  & 64 bits  & 16 bits & 32 bits  & 48 bits  & 64 bits  & 16 bits & 32 bits  & 48 bits  & 64 bits \\
        \hline
        DPH & \textbf{0.3252} & \textbf{0.4803} & \textbf{0.5279} & \textbf{0.5492} & \textbf{0.6689} & \textbf{0.7014} & \textbf{0.7218} & \textbf{0.7312} & \textbf{0.7100} & \textbf{0.7435} & \textbf{0.7544} & \textbf{0.7614} \\
        HashNet \cite{cite:ICCV17HashNet} & \underline{0.2959} & \underline{0.4211} & \underline{0.4836} & \underline{0.5056} & \underline{0.6294} & \underline{0.6604} & \underline{0.6785} & \underline{0.6883} & \underline{0.6529} & \underline{0.7025} & \underline{0.7231} & \underline{0.7335} \\
        DHN \cite{cite:AAAI16DHN} & 0.2734 & 0.3821 & 0.4352 & 0.4921 & 0.5940 & 0.6186 & 0.6277 & 0.6343 & 0.6206 & 0.6445 & 0.6641 & 0.6685 \\
        DSH \cite{cite:CVPR2016DSH} & 0.2665 & 0.3513 & 0.3963 & 0.4213 & 0.5880 & 0.6102 & 0.6155 & 0.6242 & 0.6242 & 0.6369 & 0.6481 & 0.6511 \\
        DPSH \cite{cite:IJCAI16DPSH} & 0.1625 & 0.3037 & 0.4074 & 0.4841 & 0.4041 & 0.4826 & 0.5067 & 0.5379 & 0.5660 & 0.6269 & 0.6587 & 0.6906 \\
        DNNH \cite{cite:CVPR15DNNH} & 0.1189 & 0.2783 & 0.3328 & 0.3456 & 0.3845 & 0.4632 & 0.4901 & 0.5158 & 0.5461 & 0.5923 & 0.6147 & 0.6272 \\
        CNNH \cite{cite:AAAI14CNNH} & 0.0890 & 0.2459 & 0.2903 & 0.3156 & 0.3653 & 0.4432 & 0.4672 & 0.4801 & 0.5659 & 0.5624 & 0.5519 & 0.5687 \\
        SDH \cite{cite:CVPR15SDH} & 0.1023 & 0.1543 & 0.1785 & 0.2043 & 0.3021 & 0.4056 & 0.4329 & 0.4675 & 0.4933 & 0.5323 & 0.5493 & 0.5577 \\
        KSH \cite{cite:CVPR12KSH} & 0.0823 & 0.1121 & 0.1341 & 0.1456 & 0.2007 & 0.2704 & 0.3002 & 0.3257 & 0.4946 & 0.5193 & 0.5224 & 0.5316 \\
        ITQ \cite{cite:CVPR11ITQ} & 0.2546 & 0.3313 & 0.3564 & 0.4023 & 0.3309 & 0.4179 & 0.4461 & 0.4774 & 0.5289 & 0.5824 & 0.6227 & 0.6320 \\
        SH \cite{cite:NIPS09SH} & 0.0917 & 0.1234 & 0.1524 & 0.1623 & 0.1649 & 0.1904 & 0.2115 & 0.2605 & 0.4429 & 0.4910 & 0.4782 & 0.5073 \\
        LSH \cite{cite:VLDB99LSH} & 0.0412 & 0.0623 & 0.0889 & 0.1021 & 0.0557 & 0.0938 & 0.1392 & 0.1724 & 0.3932 & 0.4659 & 0.5301 & 0.5171 \\
        \Xhline{1.0pt}
    \end{tabular}
\end{table*}

\begin{figure*}[!phtb]
    \centering
    \subfigure[Precision within Hamming radius 2]{
        \includegraphics[width=0.28\textwidth]{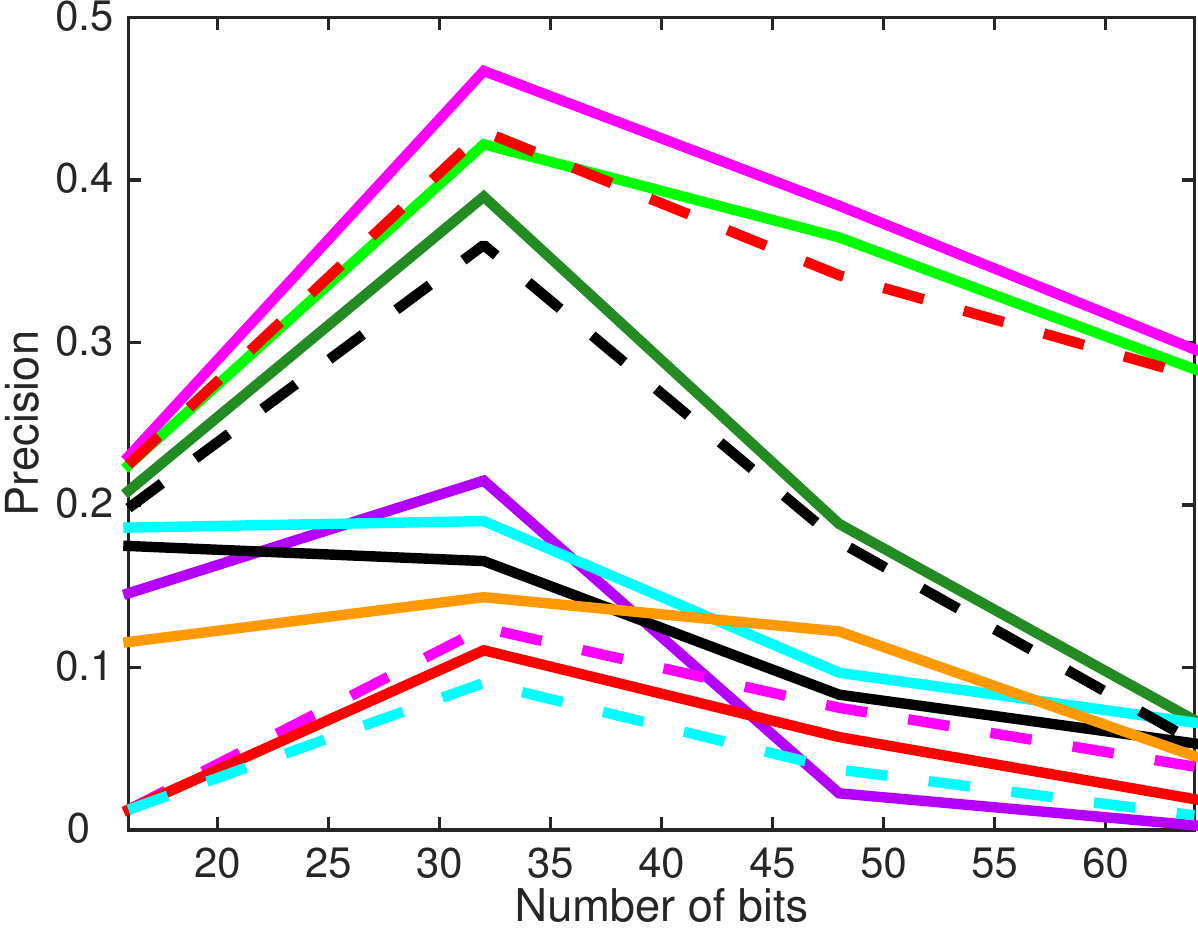}
        \label{fig:ham_imagenet}
    }
		\hfil
    \subfigure[Precision-recall curve @ 64 bits]{
        \includegraphics[width=0.28\textwidth]{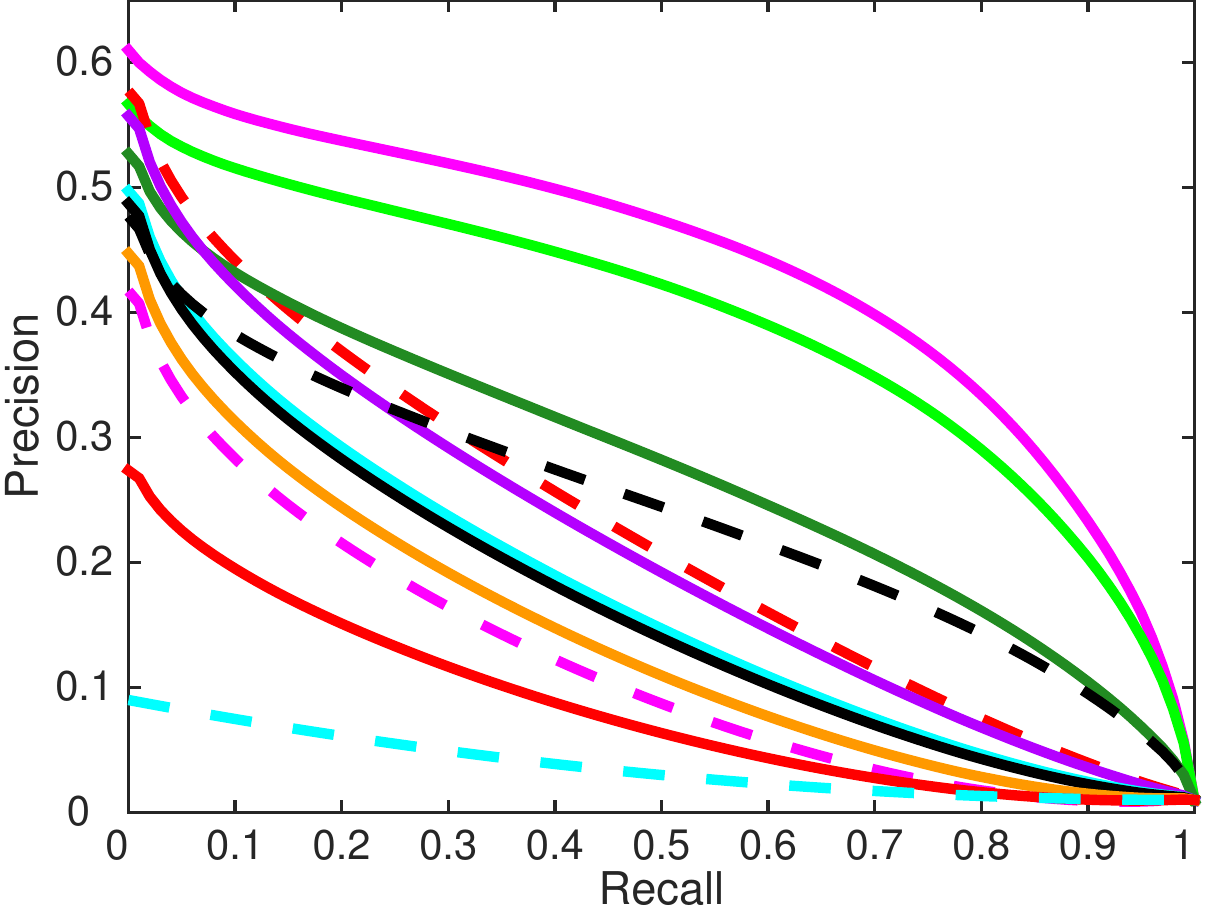}
        \label{fig:pr_imagenet}
    }
		\hfil
    \subfigure[Precision curve w.r.t. top-$N$ @ 64 bits]{
        \includegraphics[width=0.355\textwidth]{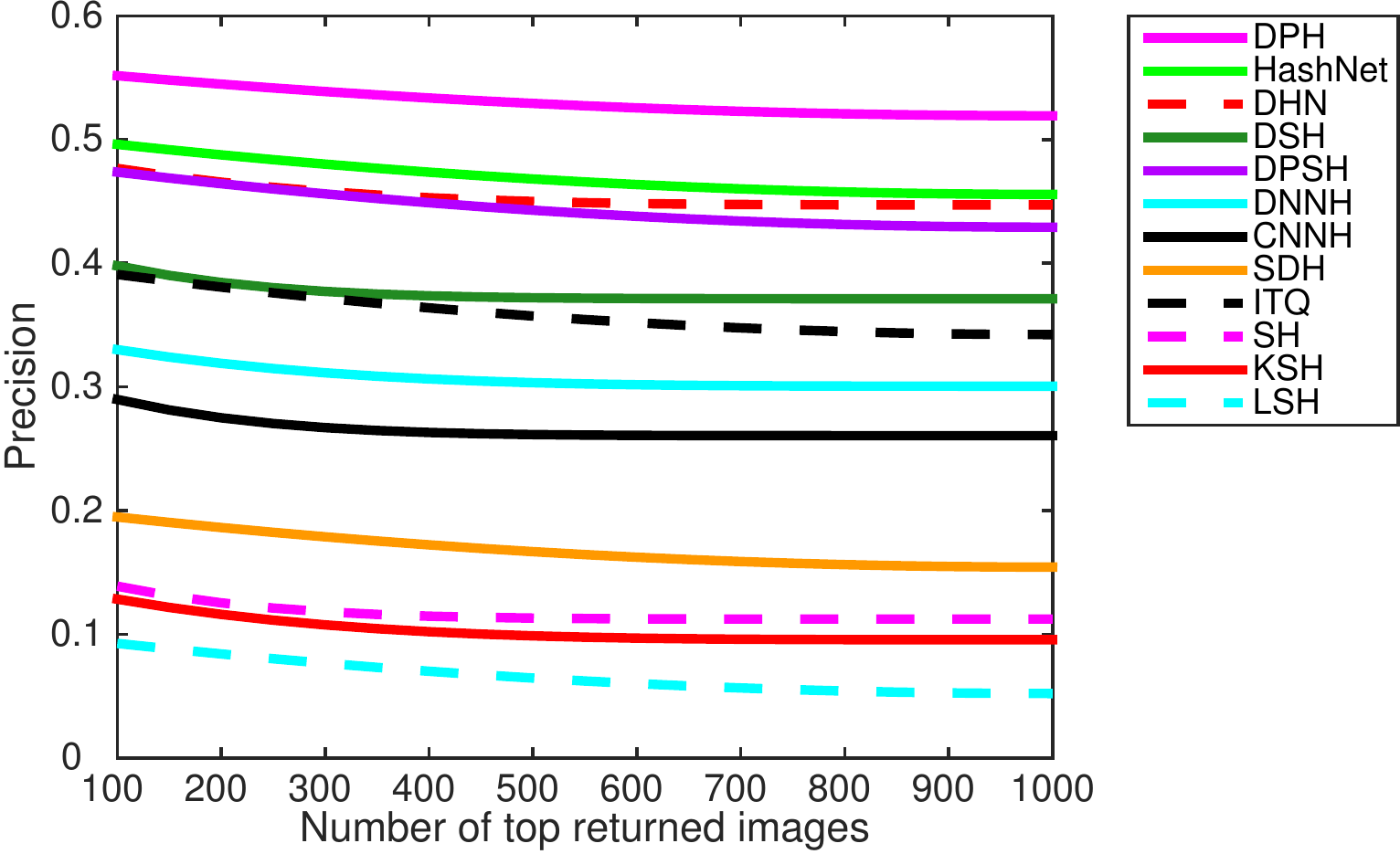}
        \label{fig:prec_imagenet}
    }
    \caption{The experimental results of DPH and comparison methods on the ImageNet dataset under three evaluation metrics.}
    \label{fig:imagenet}
\end{figure*}

\begin{figure*}[!phtb]
    \centering
    \subfigure[Precision within Hamming radius 2]{
        \includegraphics[width=0.28\textwidth]{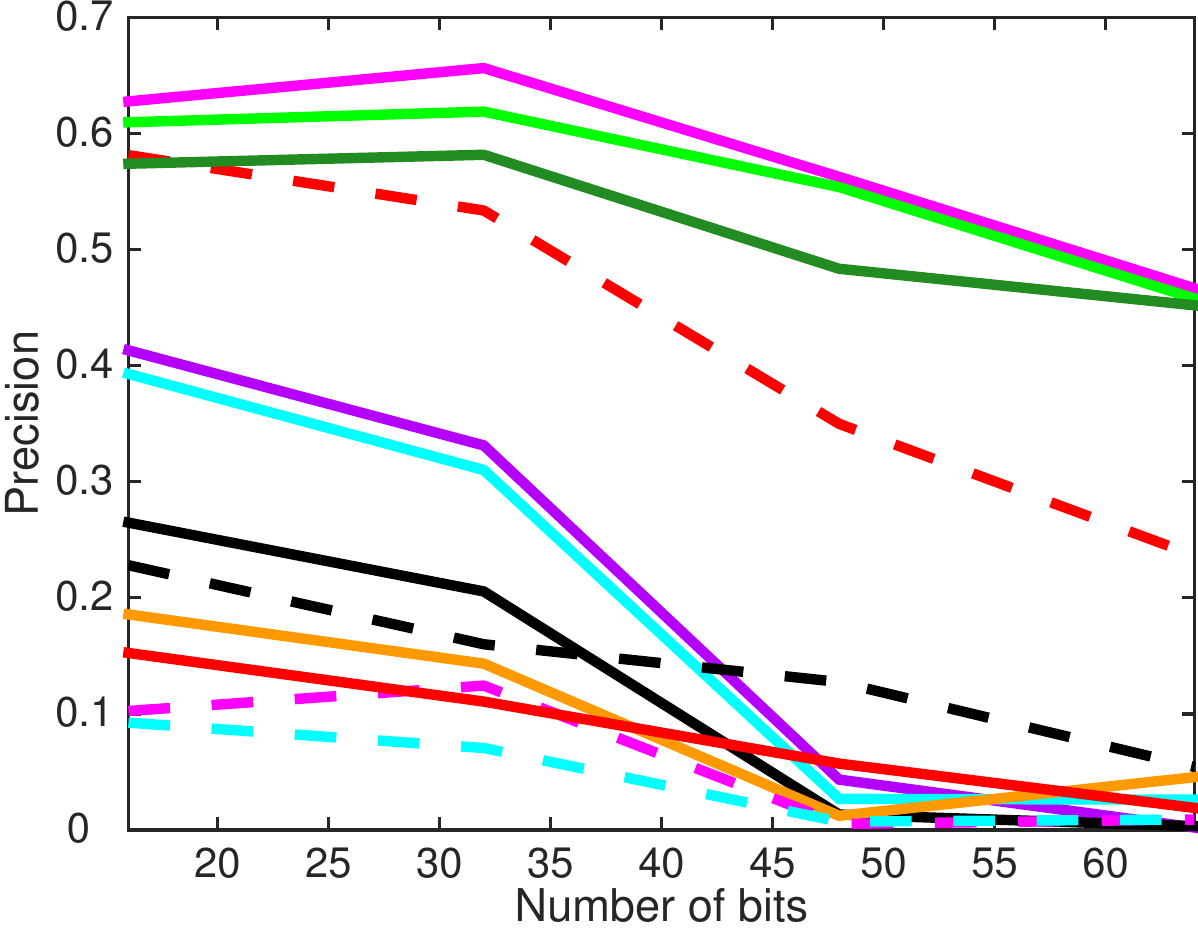}
        \label{fig:ham_nus}
    }
		\hfil
    \subfigure[Precision-recall curve @ 64 bits]{
        \includegraphics[width=0.28\textwidth]{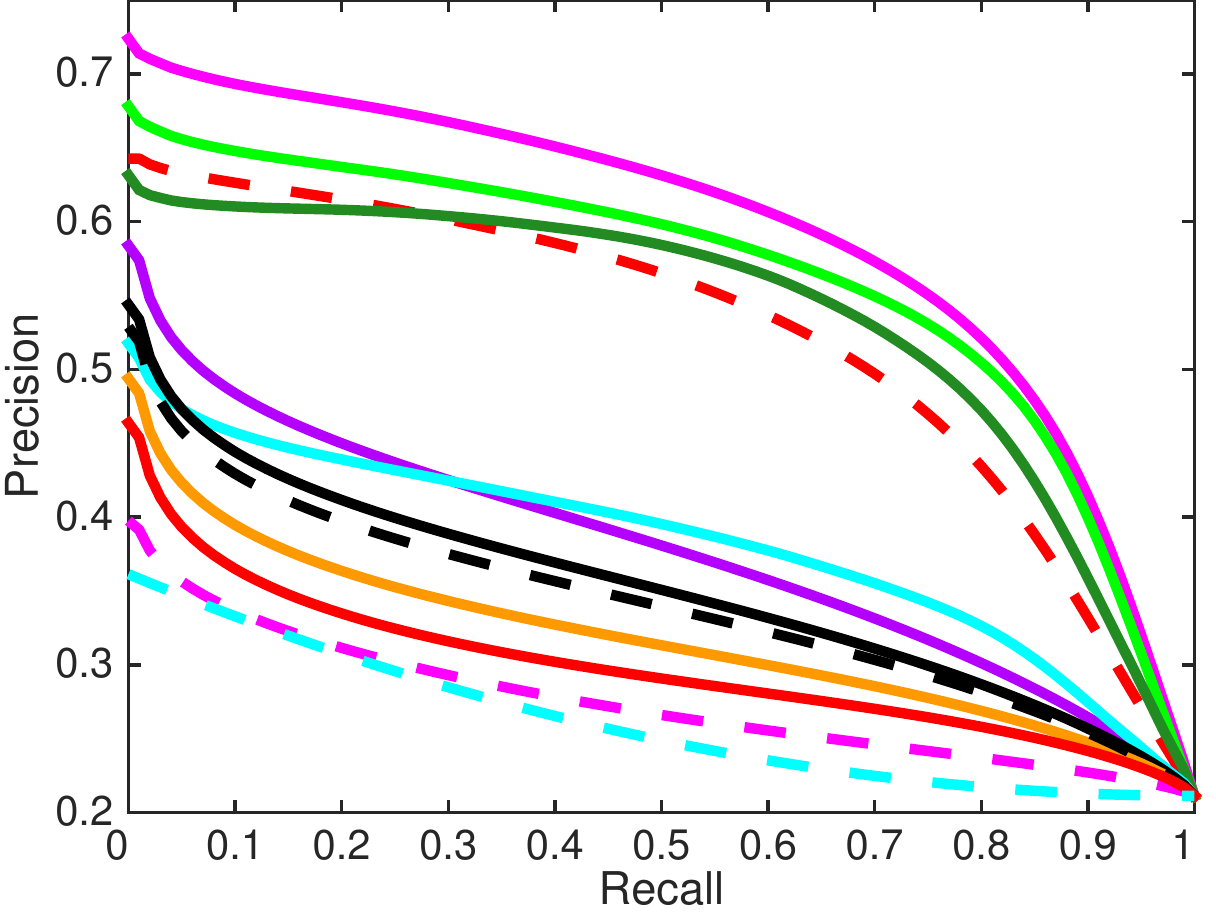}
        \label{fig:pr_nus}
    }
		\hfil
    \subfigure[Precision curve w.r.t. top-$N$ @ 64 bits]{
        \includegraphics[width=0.355\textwidth]{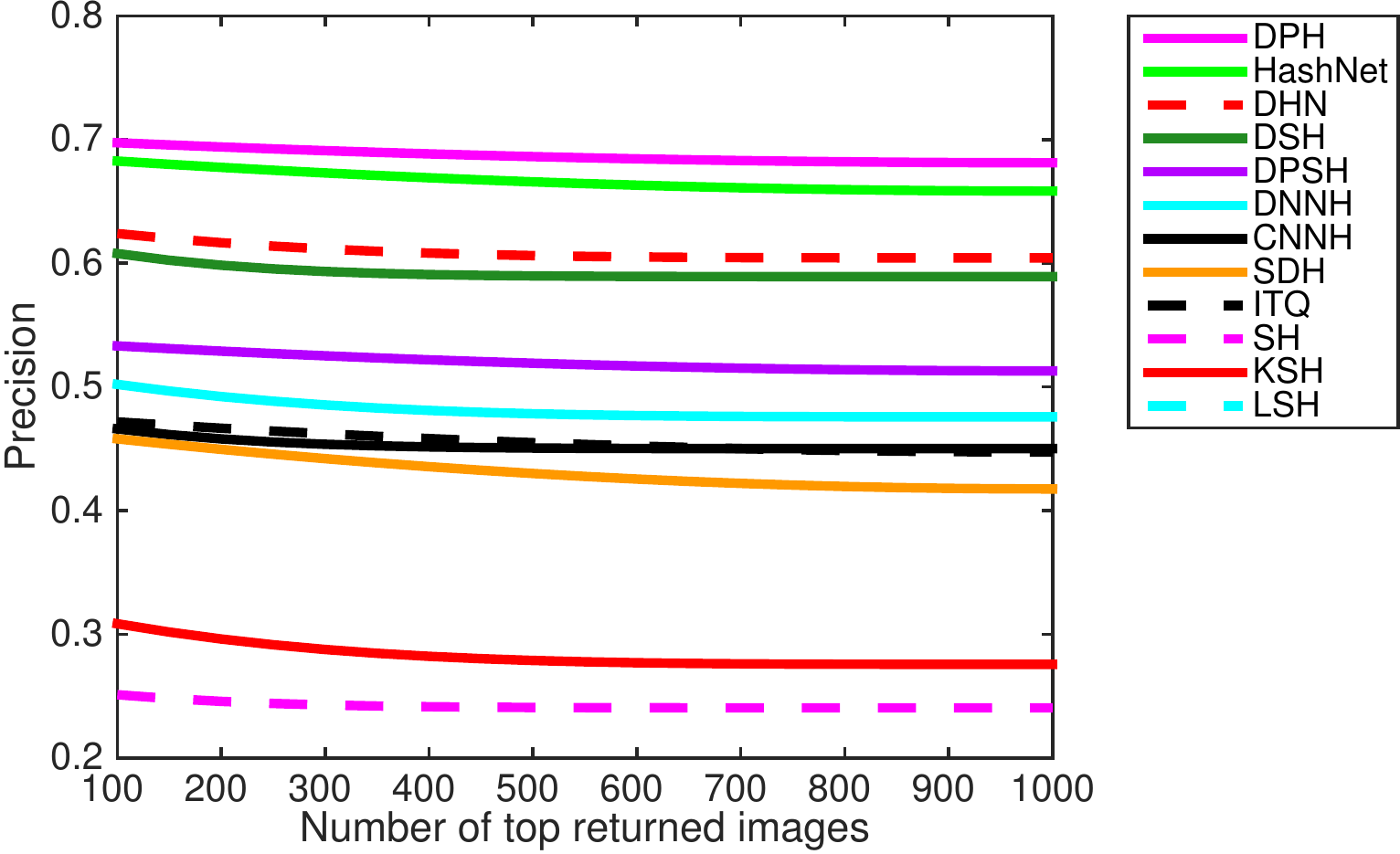}
        \label{fig:prec_nus}
    }
    \caption{The experimental results of DPH and comparison methods on the NUS-WIDE dataset under three evaluation metrics.}
    \label{fig:nus}
\end{figure*}

\begin{figure*}[!phtb]
    \centering
    \subfigure[Precision within Hamming radius 2]{
        \includegraphics[width=0.28\textwidth]{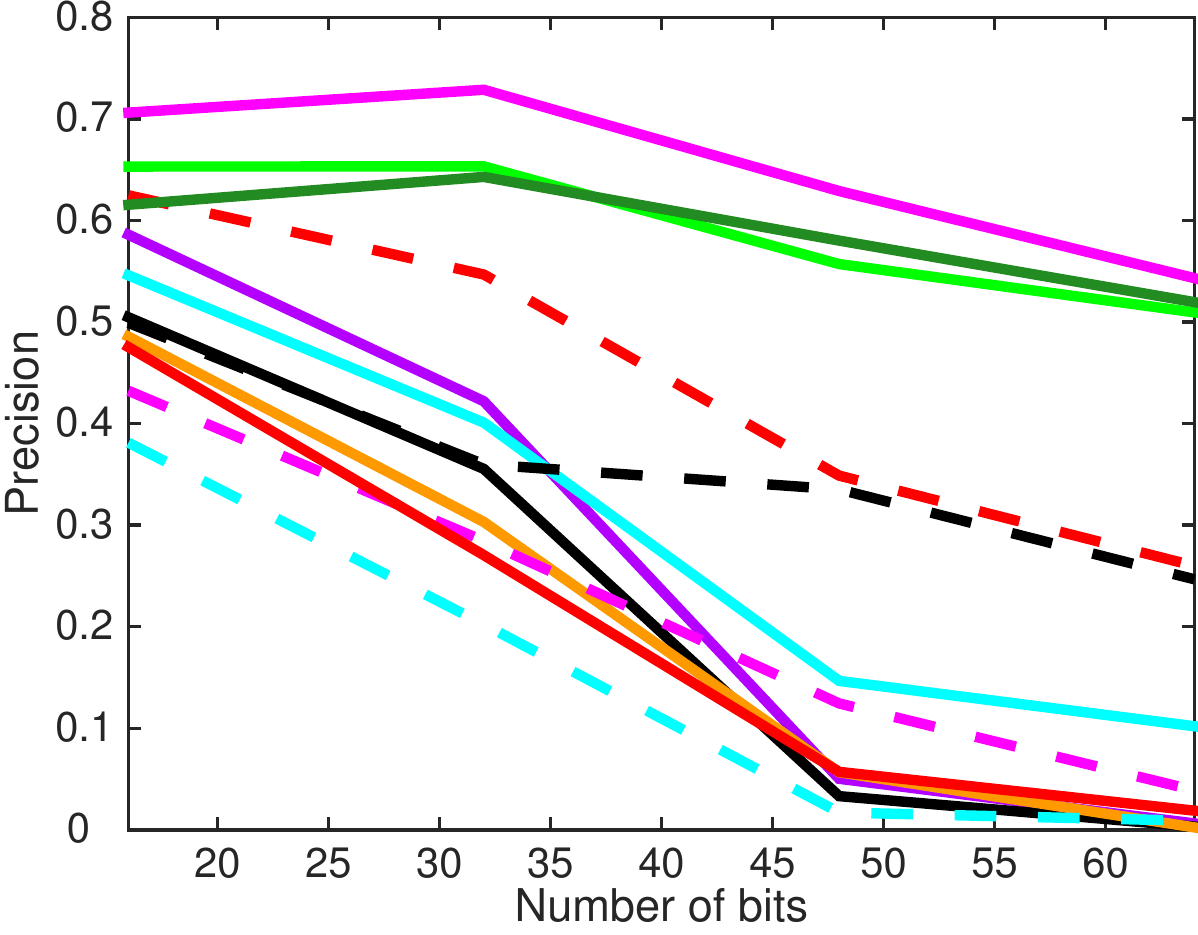}
        \label{fig:ham_coco}
    }
		\hfil
    \subfigure[Precision-recall curve @ 64 bits]{
        \includegraphics[width=0.28\textwidth]{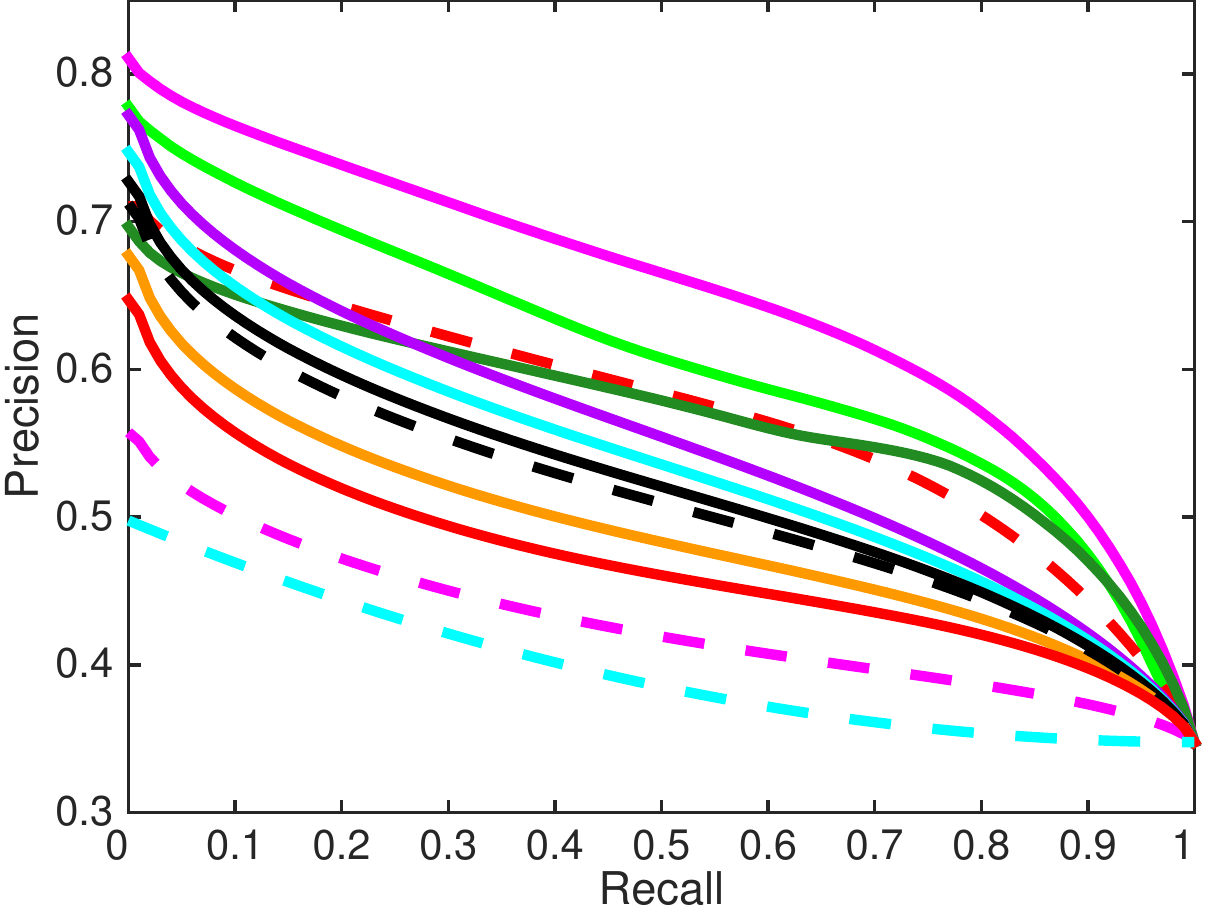}
        \label{fig:pr_coco}
    }
		\hfil
    \subfigure[Precision curve w.r.t. top-$N$ @ 64 bits]{
        \includegraphics[width=0.355\textwidth]{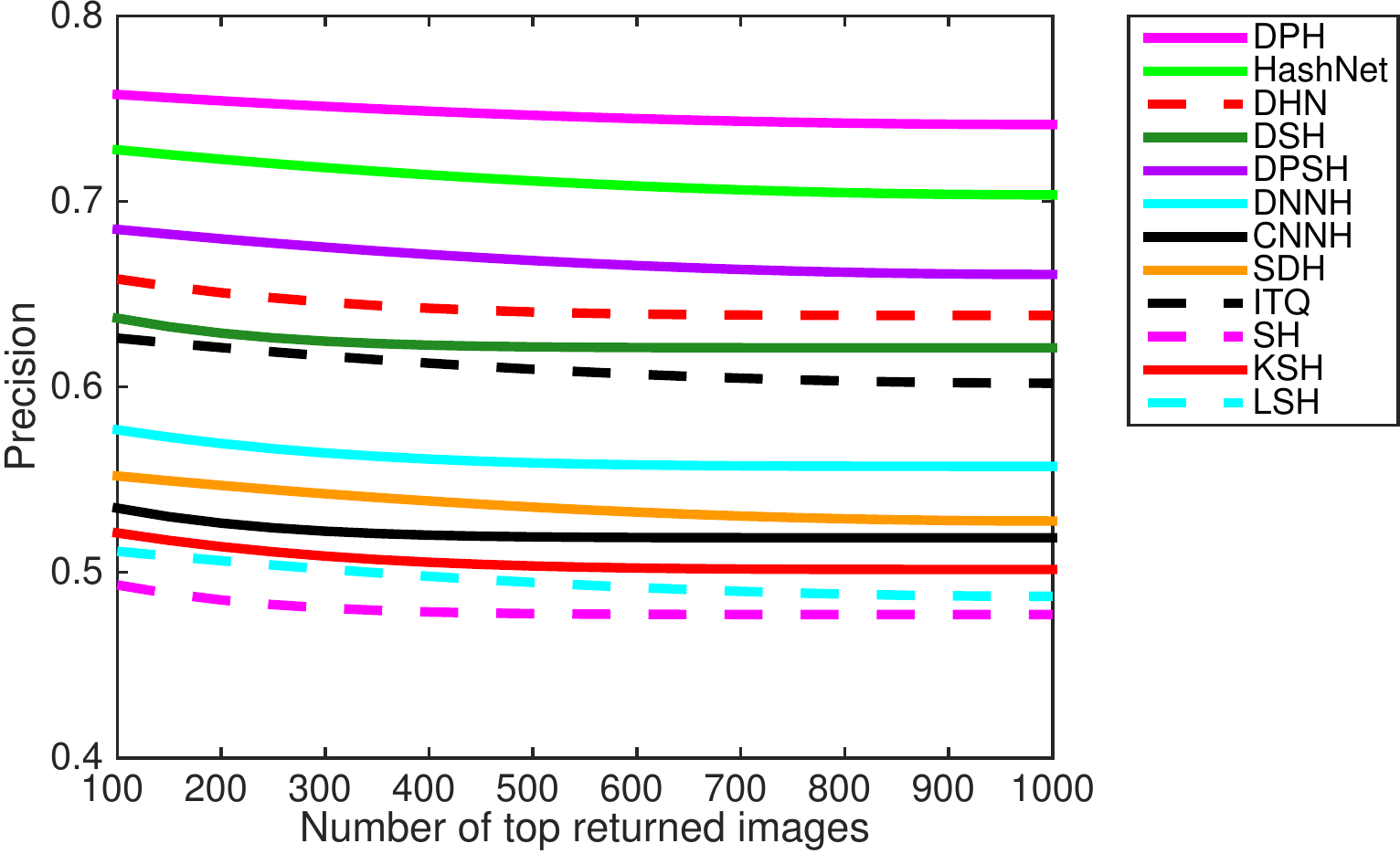}
        \label{fig:prec_coco}
    }
    \caption{The experimental results of DPH and comparison methods on the MS-COCO dataset under three evaluation metrics.}
    \label{fig:coco}
\end{figure*}

We evaluate the retrieval quality on four standard metrics:  Mean Average Precision (\textbf{MAP}), Precision-Recall curves (\textbf{PR}), Precision curves within Hamming distance 2 (\textbf{P@H$\le$2}), and Precision curves with respect to different numbers of top returned samples (\textbf{P@N}). For fair comparison, all methods use identical training and test sets. We adopt MAP@1000 for ImageNet and MAP@5000 for the other datasets as in \cite{cite:ICCV17HashNet}. It is worth noting that, in order to evaluate the retrieval performance equally on large and small classes, we let each class appears nearly equally in the query set for each dataset. This is a reasonable setting under data skewness.
 
For shallow hashing methods, we use DeCAF$_7$ features \cite{cite:ICML14DeCAF}. For deep hashing methods, we use raw images as input. We use AlexNet \cite{cite:NIPS12CNN} for all deep hashing methods, and implement DPH in \textbf{Caffe} \cite{cite:MM14Caffe}. We fine-tune convolutional layers and fully-connected layers \textit{conv1}--\textit{fc7} copied from AlexNet pre-trained on ImageNet 2012 and train the hash layer \textit{fch} by back-propagation. As the \textit{fch} layer is trained from scratch, we set its learning rate to be 10 times that of the lower layers. We use mini-batch stochastic gradient descent (SGD) with 0.9 momentum and the learning rate annealing strategy implemented in Caffe, and cross-validate the learning rate from $10^{-5}$ to $10^{-3}$ with a multiplicative step-size ${10}^{\frac{1}{2}}$. We fix the mini-batch size of images as $256$ and the weight decay parameter as $0.0005$. We select the hyper-parameters of all methods through three-fold cross-valuation.

\begin{table*}
    \caption{MAP on the Balanced ImageNet Dataset (\textbf{left}) and on the Three Imbalanced Image Datasets with VGG-16 (\textbf{right})}
    \label{fig:balanced_vgg}
    \begin{tabular}{c|cccc}
        \Xhline{1.0pt}
        \multirow{2}{30pt}{\centering Method} & \multicolumn{4}{c}{ImageNet} \\
        \cline{2-5}
        & 16 bits & 32 bits  & 48 bits  & 64 bits \\
        \hline
        DPH & \textbf{0.5168} & \textbf{0.6409} & \textbf{0.6723} & \textbf{0.6967} \\
        HashNet \cite{cite:ICCV17HashNet} & \underline{0.5059} & \underline{0.6306} & \underline{0.6633} & \underline{0.6835} \\
        \Xhline{1.0pt}
    \end{tabular}
    \hfil
    \begin{tabular}{c|cc|cc|cc}
        \Xhline{1.0pt}
        \multirow{2}{30pt}{\centering Method} & \multicolumn{2}{c|}{ImageNet} & \multicolumn{2}{c|}{NUS-WIDE} & \multicolumn{2}{c}{MS-COCO} \\
        \cline{2-7}
        & 16 bits & 64 bits  & 16 bits & 64 bits  & 16 bits & 64 bits \\
        \hline
        DPH & \textbf{0.3455} & \textbf{0.6395} & \textbf{0.7107} & \textbf{0.7439} & \textbf{0.7252} & \textbf{0.7958} \\
        HashNet \cite{cite:ICCV17HashNet} & \underline{0.3252} & \underline{0.5891} & \underline{0.6853} & \underline{0.7163} & \underline{0.6835} & \underline{0.7581} \\
        \Xhline{1.0pt}
    \end{tabular}
\end{table*}

\subsection{Results}

Table \ref{table:MAP} shows the \textbf{MAP} results. DPH outperforms all comparison methods substantially. 
Compared to ITQ, the best shallow hashing method using deep features, we achieve absolute boosts of $13.45\%$, $28.78\%$, and $15.08\%$ in average MAP for different bits on ImageNet, NUS-WIDE, and MS-COCO, respectively. Compared to HashNet, the state-of-the-art deep hashing method, we achieve absolute boosts of $4.41\%$, $4.17\%$, $3.93\%$ in average MAP for different bits on the three datasets, respectively.

HashNet uses Weighted Maximum Likelihood to tackle the data imbalance problem but it only uses the weight between positive and negative pairs, which cannot capture the the more fine-grained class imbalance issue. This class imbalance issue substantially deteriorates the average performance on all the classes since previous hashing methods will unavoidably focus on large classes and underperform on small classes. Thus, with each class appears nearly equally in the query set, the results on each dataset are worse than those reported by the original paper. However, our DPH puts larger weights on the samples of small classes to focus more on difficult examples (usually samples of small classes). Thus, DPH performs relatively well on all the classes and yields the best performance.

The performance in terms of Precision within Hamming radius 2 (\textbf{P@H=2}) is important for efficient retrieval with  hash codes since such Hamming ranking only requires $O1)$ time for each query. As shown in Figures \ref{fig:ham_imagenet}, \ref{fig:ham_nus} and \ref{fig:ham_coco}, DPH achieves the highest P@H=2 results on all three datasets. In particular, P@H=2 of DPH with 32 bits is better than that of HashNet with any bits. This validates that DPH can learn more compact hash codes than HashNet. When using longer codes, the Hamming space will become sparse and few data points fall within the Hamming ball with radius 2 \cite{cite:CVPR12MIH}. This is why most hashing methods achieve best accuracy with relatively shorter code lengths.

The retrieval performance on the three datasets in terms of Precision-Recall curves (\textbf{PR}) and Precision curves with respect to different numbers of top returned samples (\textbf{P@N}) are shown in Figures \ref{fig:pr_imagenet}$\sim$\ref{fig:pr_coco} and Figures \ref{fig:prec_imagenet}$\sim$\ref{fig:prec_coco}, respectively. DPH outperforms comparison methods by large margins. In particular, DPH achieves much higher precision at lower recall levels or when the number of top results is small. This is desirable for precision-first retrieval, which is widely implemented in practical systems.

To enable direct comparison with published papers, we also test our method on the Balanced ImageNet dataset as in HashNet~\cite{cite:ICCV17HashNet}. As shown in the left part of Table~\ref{fig:balanced_vgg}, our DPH outperforms HashNet even on the balanced ImageNet dataset, which proves that our method can perform well even on balanced dataset.

As the network structure can influence the performance of general-purpose computer vision models, we test our model on the three datasets with VGG-16 Net \cite{cite:ICLR15VGG}. Observing from the right part of Table~\ref{fig:balanced_vgg}, our DPH still outperforms HashNet, which demonstrates that our method is robust to the base network architecture.

\begin{table*}[htb]
    \centering
    \addtolength{\tabcolsep}{0pt}
    \caption{MAP Results of DPH and Its Variants, DPH-F, DPH-W, and DPH-Q on the Three Image Datasets}
    \label{table:ablation}
    \begin{tabular}{c|cccc|cccc|cccc}
        \Xhline{1.0pt}
        \multirow{2}{30pt}{\centering Method} & \multicolumn{4}{c|}{ImageNet} & \multicolumn{4}{c|}{NUS-WIDE} & \multicolumn{4}{c}{MS-COCO} \\
        \cline{2-13}
        & 16 bits & 32 bits  & 48 bits  & 64 bits & 16 bits & 32 bits  & 48 bits  & 64 bits & 16 bits & 32 bits  & 48 bits  & 64 bits \\
        \hline
        DPH & \textbf{0.3252} & \textbf{0.4803} & \textbf{0.5279} & \textbf{0.5592} & \underline{0.6689} & \underline{0.7014} & \underline{0.7218} & \underline{0.7312} & \underline{0.7100} & \textbf{0.7435} & \textbf{0.7544} & \textbf{0.7614} \\
        DPH-F & \underline{0.3115} & \underline{0.4683} & 0.5139 & \underline{0.5487} & 0.6451 & 0.6798 & 0.6918 & 0.7012 & \underline{0.7100} & \underline{0.7335} & 0.7434 & \underline{0.7555} \\
        DPH-W & 0.1919 & 0.3645 & 0.4473 & 0.4767 & 0.6210 & 0.6535 & 0.6652 & 0.6703 & 0.6826 & 0.7180 & 0.7314 & 0.7356  \\
        DPH-Q & 0.3102 & 0.4656 & \underline{0.5162} & 0.5431 & 0.6132 & 0.6557 & 0.6723 & 0.6898 & 0.6978 & 0.7329 & \underline{0.7475} & \underline{0.7555} \\
        \Xhline{1.0pt}
    \end{tabular}
\end{table*}

\subsection{Discussion}

\subsubsection{Ablation Study}
We dive deeper into the efficacy of our DPH model. We study three variants of DPH: {1)} \textbf{DPH-F}, replace the modulating factor to that of the focal loss, i.e. using $p$ instead of $q$ in the modulating factor; {2)} \textbf{DPH-W}, variant without priority weight, i.e. $w_{ij} = 1$; {3)} \textbf{DPH-Q}, variant without priority quantization loss, i.e. $\epsilon = \infty$.
We compare these variants in Table~\ref{table:ablation}.

As expected, DPH substantially outperforms DPH-W by large margins of $10.31\%$, $5.33\%$ and $2.54\%$ in average MAP for different bits on ImageNet, NUS-WIDE and MS-COCO, respectively. The classical pairwise cross-entropy loss (without priority weighting) has been widely adopted in previous work \cite{cite:AAAI14CNNH,cite:AAAI16DHN}. However, this classical loss does not account for the class imbalance and for the variations in easy and hard examples. Thus it may suffer from performance drop when training data is highly skew (e.g. NUS-WIDE) or has large variations of easy and hard examples. In contrast, our DPH model uses the proposed priority cross-entropy loss, which is a principled solution to these data skewness problems.

Furthermore, we notice that DPH outperforms DPH-F, which demonstrates the sub-optimality in using classification uncertainty $p$ in the focal loss. Our priority weight can relax the restriction on the choices of the difficulty measure in the modulating factor. 

DPH outperforms DPH-Q by 1.44\%, 4.81\%, and 0.89\%  in average MAP for different code lengths on ImageNet, NUS-WIDE, and MS-COCO respectively. These results validate that the proposed priority quantization loss \eqref{eqn:Q} can control the quantization error caused by continuous relaxation and generate less lossy binary codes.

\begin{figure}[h]
    \includegraphics[width=1\columnwidth]{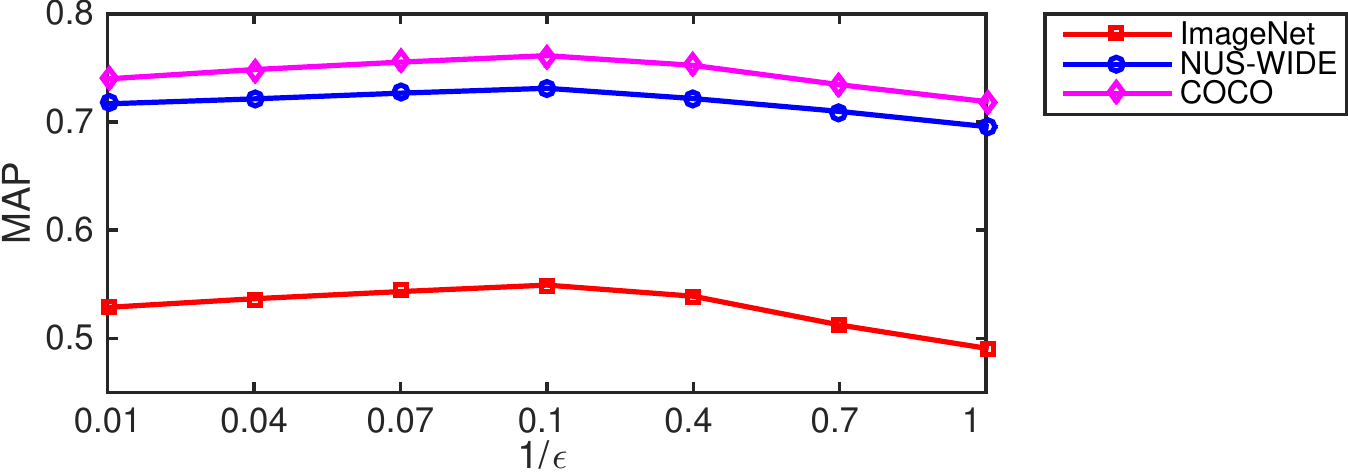}
    \caption{Sensitivity of $\frac{1}{\epsilon}$ for DPH on the three datasets.}
    \label{fig:parameter}
\end{figure}

\subsubsection{Parameter Sensitivity}
We investigate the sensitivity of the tradeoff hyper-parameter $\frac{1}{\epsilon}$ between the Priority Cross-Entropy Loss and the Priority quantization Loss on ImageNet, NUS-WIDE and MS-COCO datasets. From Figure~\ref{fig:parameter}, we can observe that the MAP results fluctuate slightly as $\frac{1}{\epsilon}$ increases from $0.01$ to $1$. This demonstrates that DPH is not sensitive to the scale of the tradeoff hyper-parameter and can perform stably in a wide range of $\frac{1}{\epsilon}$.

\begin{figure}[h]
    \centering
    \subfigure[DPH]{
        \includegraphics[width=0.45\columnwidth]{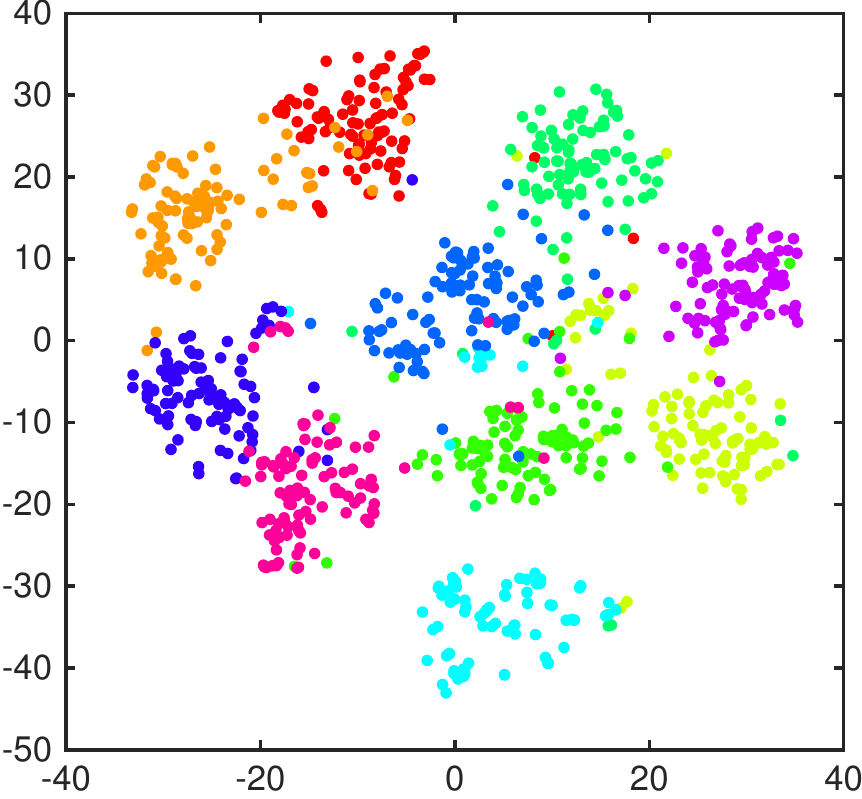}
        \label{fig:t-sne_focal}
    }
    \hfil
    \subfigure[HashNet]{
        \includegraphics[width=0.44\columnwidth]{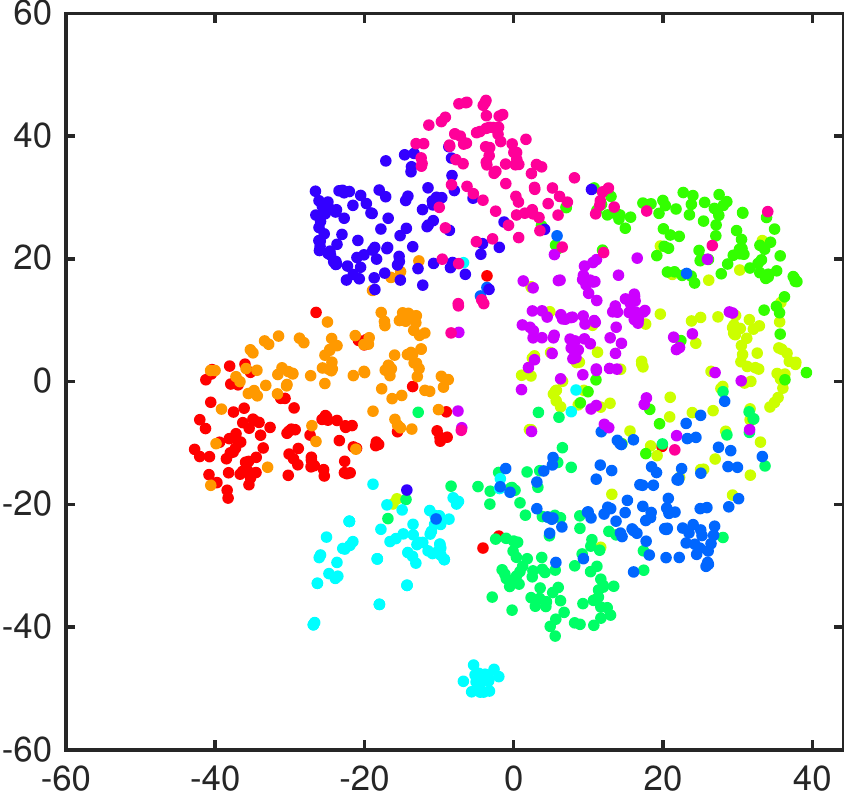}
        \label{fig:t-sne_hashnet}
    }
    \caption{t-SNE of hash codes learned by DPH and HashNet.}
    \label{fig:t-sne}
\end{figure}

\begin{figure}[!tbp]
  \centering
  \includegraphics[width=1.0\columnwidth]{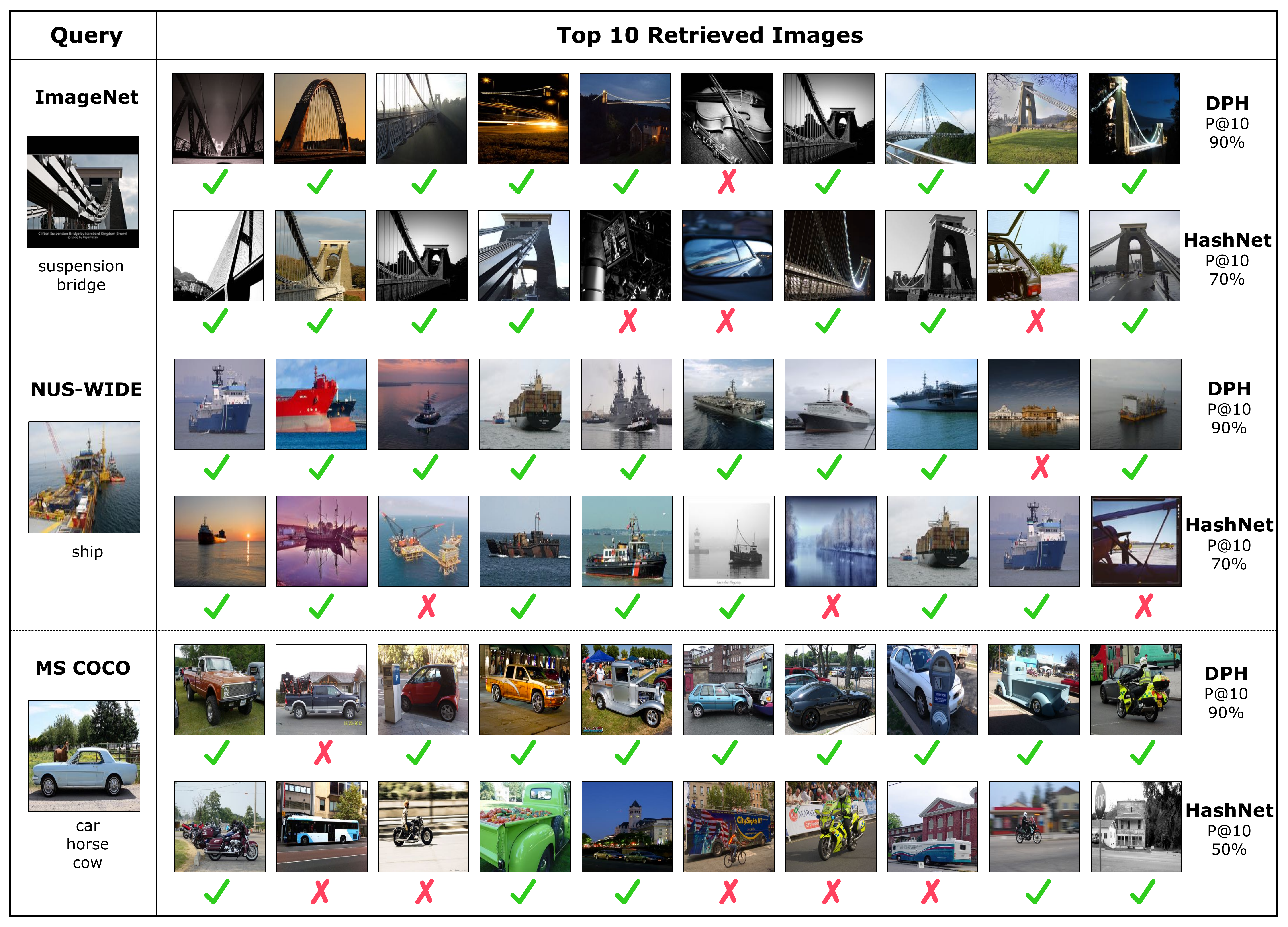}
  \caption{Examples of the top 10 images and Precision@10.}
   \label{fig:top10}
\end{figure}

\subsubsection{Visualization}

We visualize the t-SNE \cite{cite:ICML14DeCAF} of hash codes generated by HashNet and DPH on ImageNet in Figure~\ref{fig:t-sne} (we sample 10 categories for ease of visualization). We observe that the hash codes generated by DPH show clear discriminative structures in that different categories are well separated, while those generated by HashNet do not show such discriminative structures. 

Figure \ref{fig:top10} illustrates the top 10 returned images of DPH and the best deep hashing baseline HashNet \cite{cite:ICCV17HashNet} for three query images on the three datasets ImageNet, NUS-WIDE, and MS-COCO, respectively. DPH yields much more relevant and user-desired retrieval results than the state-of-the-art method.

\section{Conclusion}

This paper studies deep learning to hash approaches to establish efficient and effective image retrieval under a variety of data skewness scenarios.
The proposed Deep Priority Hashing (DPH) approach generates compact and balanced hash codes by jointly optimizing a novel priority cross-entropy loss and a priority quantization loss in a single Bayesian learning framework.
The overall model can be trained end-to-end with well-specified loss functions.
Extensive experiments demonstrate that DPH can yield state-of-the-art image retrieval performance under skewness on three benchmark datasets, ImageNet, NUS-WIDE, and MS-COCO.

In the future, we plan to extend the probabilistic framework to support image retrieval with relative similarity information, i.e. there is only similarity information on whether an image is more similar to another image than a third image.

\section{Acknowledgements}

This work is supported by National Key R\&D Program of China (2016YFB1000701), and NSFC grants (61772299, 61672313, 71690231).

{
\bibliographystyle{ACM-Reference-Format}
\bibliography{DPH}
}

\end{document}